\pdfoutput=1

\documentclass[11pt]{article}

\usepackage[final]{acl}

\usepackage{times}
\usepackage{latexsym}

\usepackage[T1]{fontenc}
\usepackage{hyperref}
\usepackage{graphicx}
\usepackage{url}
\usepackage{listings}
\usepackage{amsmath}
\usepackage{dsfont}
\usepackage{booktabs}
\usepackage{color}
\usepackage{subcaption}
\usepackage[utf8]{inputenc}
\usepackage{amsfonts}
\usepackage{tabularx}

\usepackage{microtype}

\usepackage{inconsolata}

\usepackage{graphicx}

\title{Improving the Language Understanding Capabilities of Large Language Models Using Reinforcement Learning}

\author{
    Sai Ashish Somayajula\footnotemark \quad Bokai Hu\footnotemark[1] \quad Qi Cao \quad Xin Pan \quad Pengtao Xie \\
    UC San Diego\\
    \texttt{\{ssomayaj, boh001, q9cao, p1xie\}@ucsd.edu}
}

\begin{document}
\maketitle
\renewcommand{\thefootnote}{\fnsymbol{footnote}}
\footnotetext[1]{Equal Contribution}
\renewcommand{\thefootnote}{\arabic{footnote}}
\setcounter{footnote}{0}

\begin{abstract}

Instruction-fine-tuned large language models (LLMs) under 14B parameters continue to underperform on natural language understanding (NLU) tasks, often trailing smaller models like BERT-base on benchmarks such as GLUE and SuperGLUE. Motivated by the success of reinforcement learning in reasoning tasks (e.g., DeepSeek), we explore Proximal Policy Optimization (PPO) as a framework to improve the NLU capabilities of LLMs. We frame NLU as a reinforcement learning environment, treating token generation as a sequence of actions and optimizing for reward signals based on alignment with ground-truth labels. PPO consistently outperforms supervised fine-tuning, yielding an average improvement of 6.3 points on GLUE, and surpasses zero-shot and few-shot prompting by 38.7 and 26.1 points, respectively. Notably, PPO-tuned models outperform GPT-4o by over 4\% on average across sentiment and natural language inference tasks, including gains of 7.3\% on the Mental Health dataset and 10.9\% on SIGA-nli. This work highlights a promising direction for adapting LLMs to new tasks by reframing them as reinforcement learning problems, enabling learning through simple end-task rewards rather than extensive data curation. Our code is available at \href{https://github.com/coder-qicao/RL4GLUE}{https://github.com/coder-qicao/RL4GLUE}.

\end{abstract}

\section{Introduction}

\begin{figure*}[ht]
    \centering
    \includegraphics[trim={0cm 0cm 3cm 0cm}, clip, width=\linewidth]{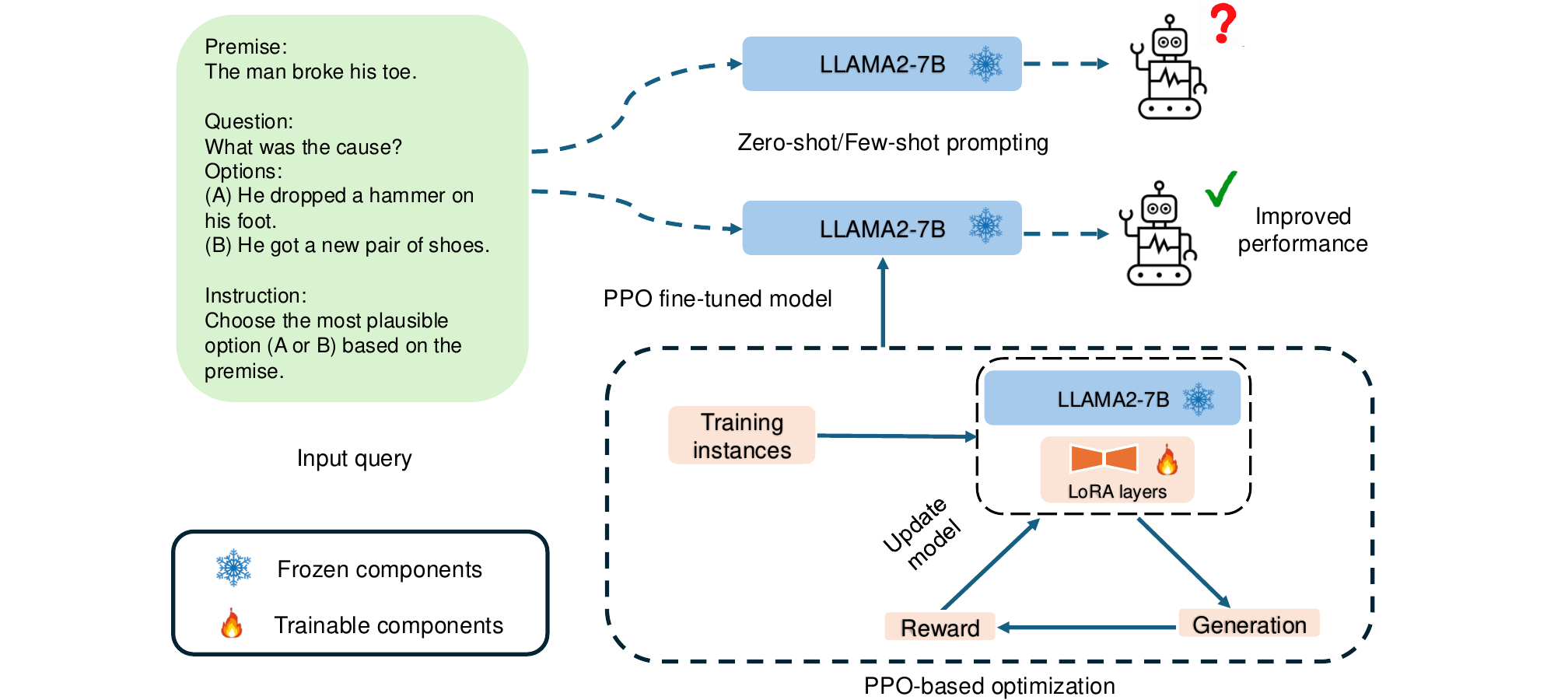}
    \caption{Despite extensive pre-training and instruction tuning, zero-shot and few-shot prompting of models like LLAMA2-7B-chat-hf and LLAMA2-13B-chat-hf continues to underperform on NLU tasks, often falling short of smaller encoder-based models like BERT. To address this performance gap, we recast the task of NLU adaptation as an RL problem, fine-tuning the model using PPO. The input consists of a prompt and a query. In the standard setting (top path), the base model, with zero-shot/few-shot prompting, struggles to generate correct answers. Our approach selectively updates lightweight LoRA layers, using PPO to optimize a reward signal based on task correctness. This optimization encourages performance gains while constraining deviation from the base policy. We find that PPO fine-tuning substantially outperforms standard SFT.}
    \label{fig:flowchart}
    \vspace{-0.25cm}
\end{figure*}

Large language models (LLMs)~\citep{radford2019language, brown2020language, touvron2023llama} have revolutionized natural language processing (NLP) with their powerful text generation capabilities~\citep{radford2018improving}. Pretrained on large-scale unlabeled corpora, LLMs can generate coherent and contextually relevant content. Using prompt-based strategies like zero-shot and few-shot prompting~\citep{brown2020language}, these models can address a wide range of downstream tasks without task-specific fine-tuning. However, when applied to instruction-fine-tuned LLMs under 14B parameters~\footnote{The 14B cutoff reflects our computational constraints; LLAMA-2-13B-chat-hf is the largest model we were able to evaluate.}—such as \textcolor{black}{LLAMA2-7B-chat-hf}—these methods often underperform on natural language understanding (NLU) tasks compared to encoder-only models like BERT~\citep{devlin2018bert}, which consistently excel on benchmarks such as GLUE~\citep{wang2018glue} and SuperGLUE~\citep{wang2020supergluestickierbenchmarkgeneralpurpose}. For instance, our evaluation of \textcolor{black}{LLAMA2-7B-chat-hf} shows that zero-shot prompting with task-specific prompts yields an average performance of 46.1 across all GLUE datasets, while few-shot prompting improves performance to 58.7—both significantly trailing BERT-base’s 79.6, as shown in Table~\ref{GLUE_result}.

To enhance NLU capabilities of LLMs, we investigate reinforcement learning (RL)-based fine-tuning approaches. Motivated by recent work such as DeepSeek~\cite{liu2024deepseek}, which demonstrates the utility of reward-driven optimization for improving reasoning abilities, we explore the use of Proximal Policy Optimization (PPO)~\citep{schulman2017proximal} to align model outputs with task-specific objectives.

While standard fine-tuning (SFT) is commonly used to adapt LLMs to downstream tasks, we find it insufficient for NLU—often underperforming even smaller encoder-only models like BERT-base. In contrast, we use PPO to enhance LLM performance by framing NLU as a reinforcement learning problem, as shown in Fig~\ref{fig:flowchart}. The sequence of input tokens up to timestep $t-1$ represents the state $s_t$, and the token generated at timestep $t$ is treated as the action $a_t$. After generating the full response, a heuristic extracts the predicted answer, which is compared to the ground-truth label to assign a scalar reward $R$. PPO then updates the model to maximize this reward, enabling direct optimization for task-specific objectives. Empirically, PPO-based fine-tuning of \textcolor{black}{LLAMA2-7B-chat-hf} improves GLUE performance by 6.3 points over SFT, surpasses zero and few-shot prompting by 38.7 and 26.1 points, respectively, and even outperforms GPT-4o by over 4\% on sentiment and inference tasks—achieving gains of 7.3\% on the Mental Health dataset and 10.8\% on SIGA-nli. These results demonstrate the effectiveness of reinforcement learning in aligning LLMs under 14B parameters with NLU objectives.


Pre-trained LLMs possess broad linguistic knowledge, spanning syntactic and semantic structures, acquired from large-scale text corpora. We show that reinforcement learning, specifically PPO, can refine this general understanding to better align with task-specific NLU objectives. Similar patterns are observed in DeepSeek, where chain-of-thought pre-training enhances reasoning capabilities, and subsequent RL-based fine-tuning further improves performance. Building on this insight, our findings suggest a promising direction: adapting LLMs to new tasks by formulating them as reinforcement learning problems. When models are sufficiently pre-trained, task alignment may be achieved without additional labeled data—requiring only a well-defined reward function over the outputs. PPO can then optimize the model toward high-reward behaviors. This approach offers a scalable, label-efficient alternative to conventional supervised fine-tuning through reward-driven adaptation.

\section{Method}
To enhance the performance of LLMs on NLU tasks, we adopt two distinct fine-tuning methods. The first approach involves supervised fine-tuning, where the input consists of a concatenation of the task-specific prompt, query and the ground truth answer, with the model optimized using the next-token prediction objective. The second approach utilizes PPO, framing response generation as a reinforcement learning problem. In this setup, the sequence of input tokens until timestep \(t-1\) represents the state \(s_t\), and each token generated at timestep \(t\) is treated as an action \(a_t\). After generating the entire sequence, a heuristic-based process extracts the final answer from this generated sequence, and is compared to the ground truth. PPO is then employed to optimize the model by maximizing the cumulative reward derived from this comparison. To reduce computational complexity, we fine-tune LoRA layers instead of the full model. We refer the readers to Appendix~\ref{sec: related-works} and \ref{sec:preliminaries} for preliminaries on PPO.

\subsection{Task-Specific Prompt Design}
\label{sec:task-specific-prompt}
We detail the construction of task-specific prompts used to query the LLM for NLU tasks. Each prompt begins with a clear task description, outlining the necessary background information to guide the model in solving the task. Following this, we specify strict requirements for the output format, ensuring that the response is encapsulated within a predefined structure, specifically between `<Judgement></Judgement>' tags. This structure ensures consistency in the model's responses, facilitating easier extraction and evaluation of the results.

For example, in the CoLA task, which assesses grammatical acceptability, the prompt is structured as follows:

\lstset{
    basicstyle=\ttfamily\footnotesize,    
    breaklines=true,                      
    frame=single,                         
    tabsize=4,                            
    columns=fixed,                        
    keepspaces=true                       
}

\begin{lstlisting}
System_prompt:
    You are an assistant to analyze the linguistic properties 
    of a sentence. The task is to decide the linguistic acceptability 
    of a sentence. If the sentence is linguistically correct then it 
    is acceptable, else it is not.
The result you give should have the following form:
    <Judgement> {Insert only "Yes" or "No" here} </Judgement>
Prompt:
    Now judge if the sentence "{sentence}" is linguistically acceptable.
Assistant:
    <Judgement>
\end{lstlisting}

The prompt starts with background information about CoLA, specifies restrictions on the output (such as labeling a sentence as acceptable or unacceptable), and concludes with a special start token, <Judgement>, to initiate the model’s response generation.

\subsection{Supervised Fine-tuning of LLM on NLU Tasks}
Given an NLU training dataset, \(\mathcal{D}^{(tr)} = \{(x_i, y_i)\}_{i=1}^N\), where \(x_i\) represents the input text and \(y_i\) the ground truth label, we fine-tune the LLM on a sequence consisting of the task-specific prompt \(p\) (described in section \ref{sec:task-specific-prompt}) concatenated with the input \(x_i\) and the ground truth answer \(y_i\). The model is trained using the next-token prediction objective, where it predicts the next token in the sequence by conditioning on all preceding tokens. This objective trains the model to learn to predict the correct answer for the NLU task conditioned on the task-specific prompt and input.

\subsection{Proximal Policy Optimization for LLM Fine-tuning on NLU Tasks}

We utilize PPO to fine-tune the LLM on NLU tasks, following the training protocol outlined in Appendix~\ref{sec:preliminaries}. The reward function is specifically designed for each NLU task. In this work, we use a simple reward function, where a reward is assigned at \emph{the end of the generation} based on alignment with the ground truth labels. We use regular expression matching to extract answers from the LLMs outputs by first locating the text within the `<Judgement></Judgement>' tags. Depending on the task, we then search for task-specific keywords (such as ``yes'', ``no'', ``acceptable'', or ``not acceptable'') to identify the answer. These extracted answers are compared with the ground truth to determine the appropriate rewards.

For instance, CoLA, a classification task, answers are categorized as \textit{acceptable}, \textit{unacceptable}, or \textit{exceptional} (incorrect format). For STS-B, a regression task, the extracted answer is a floating-point number between 0 and 5. Reward per generation for classification tasks is given by \( R = \mathds{1}(\hat{y} == y_i) \), where \(\hat{y}\) is model's prediction and \(y\) is ground truth. For STS-B, a regression task, the reward for each generation is defined as $R = 2.5 - |\hat{y}_i - y_i|$, where $\hat{y}_i$ is the predicted score and $y_i$ is the ground truth. Since both scores lie in $[0, 5]$, their absolute difference ranges from 0 to 5. Subtracting this from 2.5 centers the reward around zero and bounds it within $[-2.5,\, 2.5]$, with the maximum reward achieved when $\hat{y}_i = y_i$. \emph{Incorrectly formatted responses} are penalized with a value of -1 for classification tasks and -2.5 for regression tasks, representing the largest penalties applied in each case.

\subsection{Low-Rank Adaptation}

To mitigate the computational cost of full-model fine-tuning, we employ LoRA~\citep{hu2021loralowrankadaptationlarge} during both the supervised fine-tuning and PPO stages. Instead of updating the entire model, we restrict the updates to LoRA layers, which significantly reduces the number of trainable parameters by decomposing the weight matrices into low-rank matrices.

\section{Experiments}

\subsection{Experimental Setup}
\label{sec:exp_detais}
We trained and evaluated our models on the GLUE~\citep{wang2018glue} and SuperGLUE~\citep{wang2020supergluestickierbenchmarkgeneralpurpose} benchmarks. All experiments were conducted using instruction-tuned LLAMA2-7B models~\citep{touvron2023llama2openfoundation}\footnote{\url{https://huggingface.co/daryl149/llama-2-7b-chat-hf}}. We perform both single task and multi-task fine-tuning: 1) \textit{Single-task fine-tuning}: For each subtask within GLUE and SuperGLUE, a separate task-specific LoRA module was trained independently. 2) \textit{Multi-task fine-tuning}: In the multi-task setting, datasets from different subtasks within each benchmark were combined, and a single LoRA module was trained to handle all tasks simultaneously. Please refer to Appendix~\ref{sec:hyperparameters} for detailed hyperparameter settings.

\subsection{Baselines}

We evaluated the performance of our approach against three baselines:

\begin{itemize}
    \item \textbf{Encoder-only models}: We compare our results with encoder-only transformer models, specifically BERT-base (110M parameters) and BERT-large (340M parameters)~\citep{devlin2019bertpretrainingdeepbidirectional}.
    
    \item \textbf{Zero-shot prompting:} The model is provided with task-specific prompts, as outlined in section~\ref{sec:task-specific-prompt}, along with the input query. The model is required to generate predictions solely based on these prompts and the input query, without any additional task-specific fine-tuning.
    
    \item \textbf{Few-shot prompting:} In this setting, the model is provided with both the task-specific prompt and \textcolor{black}{one to five labeled examples (which ever gave the best performance)} from the training dataset as demonstrations. These examples are provided as reference to guide the model in generating more accurate responses for the input query. Similarly, no task-specific fine-tuning is performed.
\end{itemize}

After generating a response, we applied regular expression matching to extract the relevant answer from the model's output. We directly matched task-specific keywords (like ``yes'' or ``no'') in the generated text to identify the answer. This extracted answer was then compared to the ground truth label to evaluate the model's performance.

\subsection{Results on GLUE Benchmark}

\renewcommand\arraystretch{1.2}
\begin{table*}[!htb]
\centering
\footnotesize
\begin{tabular}{lcccccc}
  \toprule[1.2pt]
  \textbf{Models} & \textbf{MNLI-m} & \textbf{MNLI-mm} & \textbf{QQP} & \textbf{QNLI} & \textbf{SST-2} & \textbf{CoLA} \\
  \midrule[1.2pt]
  \textbf{BERT-base} & 84.6 & 83.4 & \underline{71.2} & 90.5 & 93.5 & 52.1 \\
  \textbf{BERT-large} & 86.7 & 85.9 & \textbf{72.1} & 92.7 & \underline{94.9} & \textbf{60.5} \\
  \multicolumn{1}{l}{\textbf{\textcolor{black}{LLAMA2-7B-chat-hf}}} & & & & & & \\
  \textit{Zero-shot prompting} & 38.3 & 39.7 & 31.3 & 58.5 & 75.7 & 18.6 \\
  \textit{Few-shot prompting} & 62.4 & 61.7 & 30.9 & 60.7 & 84.2 & 29.0 \\
  \textit{PPO-ST} & \textbf{88.8} & \underline{88.2} & 70.5 & \underline{93.2} & \textbf{96.4} & \underline{59.9} \\
  \textit{SFT-ST} & 87.0 & 86.5 & 63.8 & \textbf{93.6} & 73.8 & 50.7 \\
  \textit{PPO-MT} & \underline{88.7} & \textbf{88.3} & 67.3 & 90.2 & 94.6 & 47.7\\
  \textit{SFT-MT} & 84.9 & 84.5 & 62.9 & 86.0 & 72.0 & 41.4\\
  \bottomrule[1.2pt]
\end{tabular}

\vspace{0.15cm} 

\footnotesize
\begin{tabular}{lcccccc}
  \toprule[1.2pt]
  \textbf{Models} & \textbf{STS-B} & \textbf{MRPC} & \textbf{RTE} & \textbf{WNLI} & \textbf{AX} & \textbf{Average} \\
  \midrule[1.2pt]
  \textbf{BERT-base} & 85.8 & 88.9 & 66.4 & / & / & 79.6\\
  \textbf{BERT-large} & 86.5 & \underline{89.3} & 70.1 & / & / & 82.1\\
  \multicolumn{1}{l}{\textbf{\textcolor{black}{LLAMA2-7B-chat-hf}}} & & & & & & \\
  \textit{Zero-shot prompting} & 27.5 & 66.3 & 59.3 & 44.5 & 9.2 & 46.1\\
  \textit{Few-shot prompting} & 45.5 & 80.8 & 72.9 & 51.4 & 9.2 & 58.7\\
  \textit{PPO-ST} & \underline{92.6} & \textbf{89.4} & 84.3 & \underline{74.7} & \textbf{52.7} & \textbf{84.8}\\
  \textit{SFT-ST} & 84.7 & 85.8 & 80.4 & 63.7 & \underline{45.1} & 78.5\\
  \textit{PPO-MT} & \textbf{94.7} & 86.7 & \textbf{86.9} & 66.4 &{43.4} & \underline{82.9}\\
  \textit{SFT-MT} & 85.5 & 82.6 & \underline{86.2} & \textbf{76.0} & 41.2 & 76.22\\
  \bottomrule[1.2pt]
\end{tabular}
\caption{GLUE test results are scored by the evaluation server (\href{https://gluebenchmark.com/leaderboard}{GLUE benchmark}). Average column indicates the averaged performance across all the datasets excluding the WNLI and AX datasets. F1 scores are reported for QQP and MRPC, Spearman correlations for STS-B, Matthew's correlations for CoLA, and accuracy scores for the other tasks. \textit{Zero-shot prompting} refers to prompting with task-specific prompts and an input query, while \textit{Few-shot prompting} refers to prompting with task-specific prompts, 1-5 demonstrations (chosen based on the best performance), and an input query. \textit{PPO} stands for proximal policy optimization, and \textit{SFT} refers to Supervised Fine-tuning. ``ST'' represents Single-task, while ``MT'' represents Multi-task. The \textbf{bolded} results indicate the best results, and the \underline{underlined} results indicate the second-best results.}
\vspace{-0.3cm}
\label{GLUE_result}
\end{table*}

In this section, we present our experiments on the GLUE benchmark, comparing the results with encoder-only models such as BERT~\citep{devlin2019bertpretrainingdeepbidirectional}. We use the \textcolor{black}{LLAMA2-7B-chat-hf} model as the LLM for our evaluations. The baselines include zero-shot prompting and few-shot prompting. For fine-tuning methods, we compare both supervised fine-tuning and PPO across single-task and multi-task settings. The results are summarized in Table~\ref{GLUE_result}. From the results, we make the following observations.

\textbf{First}, we observed that zero-shot prompting of the \textcolor{black}{LLAMA2-7B-chat-hf} model with task-specific prompts consistently underperformed compared to the smaller BERT-base model. \textcolor{black}{LLAMA2-7B-chat-hf} struggled notably on simpler tasks like SST-2, which only required classifying sentiment as positive or negative. This underscores the model's weak language understanding capabilities, with zero-shot prompting proving inadequate compared to BERT-base. \textbf{Second}, few-shot prompting showed improvements over the zero-shot baseline, achieving an average score of 58.7 compared to 46.1, but it still lagged significantly behind the BERT-base model’s score of 79.6. \textbf{Third}, supervised fine-tuning (SFT) using LoRA modules for each task further boosted performance, bringing it closer to BERT's level with an average score of 78.5, though still slightly behind BERT-base’s 79.6. \textbf{Fourth}, fine-tuning with PPO delivered the best results, achieving an average score of 84.6, surpassing even BERT-large’s 82.1. Moreover, zero-shot and few-shot prompting of \textcolor{black}{LLAMA2-7B-chat-hf} displayed a noticeable output imbalance, with a tendency to favor certain classes or values. In contrast, models fine-tuned with PPO showed no significant bias. \textbf{Fifth}, the total computational time for PPO is approximately \emph{1.32 times} that of SFT, indicating only a marginal increase in computational costs.

Additionally, we compared the results with multi-task training, where a \emph{single LoRA module} was trained across all datasets using both SFT and PPO to reduce time complexity. We found that SFT on individual tasks outperformed its multi-task fine-tuning counterpart. However, while PPO on multi-task training did not perform as well as PPO on single-task training, it still outperformed BERT-large in average performance, achieving a score of 82.9 compared to BERT-large’s 82.1. These results demonstrate that while single-task fine-tuning yields the best performance, multi-task training with PPO can still achieve competitive results, even surpassing state-of-the-art models like BERT-large.

\renewcommand\arraystretch{1.2}
\begin{table}[!htb]
\small
\centering
\begin{tabular}{p{3cm}p{1.5cm}p{1.5cm}}
  \toprule[1.2pt]
  \textbf{Models} & \textbf{STS-B} & \textbf{COPA}\\
  \midrule[1.2pt]
  \textbf{BERT-large} & 86.5 & 70.6 \\
  \multicolumn{1}{l}{\textbf{\textcolor{black}{LLAMA2-7B-chat-hf}}} & & \\
  \textit{Zero-shot prompting} & 27.5 & 57.0 \\
  \textit{Few-shot prompting} & 45.5 & \textcolor{black}{73.4} \\
  \textit{PPO-ST} & 92.6 & 88.6 \\
  \multicolumn{1}{l}{\textbf{\textcolor{black}{Qwen2.5-7B-Instruct}}} & & \\
  \textit{Zero-shot prompting} & 83.7 & 96.6 \\
  \textit{Few-shot prompting} & 87.0 & \textcolor{black}{96.0} \\
  \textit{PPO-ST} & 92.2 & 97.0 \\
  \multicolumn{1}{l}{\textbf{\textcolor{black}{MPT-7B-chat}}} & & \\
  \textit{Zero-shot prompting} & 19.7 & 57.4 \\
  \textit{Few-shot prompting} & 21.7 & \textcolor{black}{57.2} \\
  \textit{PPO-ST} & 89.3 & 84.0 \\
  \bottomrule[1.2pt]
\end{tabular}
\caption{Performance comparison of \textcolor{black}{LLAMA2-7B-chat-hf}, \textcolor{black}{Qwen2.5-7B-Instruct}~\citep{hui2024qwen2}, and \textcolor{black}{MPT-7B-chat}~\citep{MosaicML2023Introducing} models on the GLUE STS-B and SuperGLUE COPA tasks under zero-shot prompting, few-shot prompting, and PPO based fine-tuning. Results are sourced from the official \href{https://gluebenchmark.com/leaderboard}{GLUE benchmark} and \href{https://super.gluebenchmark.com/}{SuperGLUE benchmark} evaluation servers. For STS-B, we report Spearman correlation, and for COPA, accuracy is used as the evaluation metric.}
\label{model_comparison}
\end{table}

To assess the consistency of our findings across different models, we evaluated \textcolor{black}{Qwen2.5-7B-Instruct} and \textcolor{black}{MPT-7B-chat} alongside \textcolor{black}{LLAMA2-7B-chat-hf} on the STS-B dataset from the GLUE benchmark and the COPA dataset from the SuperGLUE benchmark. The results are summarized in Table~\ref{model_comparison}. The results confirm that PPO-based fine-tuning consistently outperforms the BERT-large model, as well as the zero-shot and few-shot prompting baselines for all LLMs, highlighting its effectiveness across different LLMs. 

\subsection{\textcolor{black}{Comparison of RL Algorithms: PPO vs. GRPO}}

\renewcommand{\arraystretch}{1.2}
\begin{table*}[!htbp]
\small
\centering
\begin{tabular}{l *{6}{c} c}
\toprule
\textbf{Algorithm} 
  & \textbf{SST-2} 
  & \textbf{MRPC} 
  & \textbf{RTE} 
  & \textbf{CoLA} 
  & \textbf{QNLI} 
  & \textbf{Avg.} 
  & \textbf{Per-Step Runtime (s)} \\
\midrule
SFT   & 73.8  & 85.8 & 80.4 & 50.7 & 93.6 & 76.9 & 4.124 \\
PPO   & 96.4  & 89.4 & 84.3 & \textbf{59.9} & \textbf{93.2} & 84.6 & 4.299 \\
GRPO  & \textbf{96.7}  & \textbf{91.2} & \textbf{88.5} & 55.2 & 93.1 & \textbf{84.9} & 5.155 \\
\bottomrule
\end{tabular}
\caption{
Comparison of SFT, PPO, and GRPO fine-tuning methods on the LLAMA2-7B-chat-hf model across five GLUE benchmark tasks (SST-2, MRPC, RTE, CoLA, QNLI), along with per-step runtime.
}
\label{tab:runtime}
\end{table*}

{\color{black}
Our objective is to improve the natural language understanding capabilities of the base (policy) model through RL fine-tuning. In this context, we compare two RL approaches: PPO and Group Relative Policy Optimization (GRPO) ~\citep{shao2024deepseekmathpushinglimitsmathematical}. PPO requires updating a separate critic model to compute value functions, which introduces (modest) additional memory constraints. On the other hand, GRPO was designed to bypass the critic model entirely. Instead, GRPO samples multiple trajectories per prompt and computes each trajectory’s advantage by comparing its reward to the batch’s average (and standard deviation). This method not only simplifies the architecture but also reduces memory usage. 

For our experiments, we utilized the TRL library ~\citep{vonwerra2022trl} on a single Nvidia A100 GPU, with a batch size of 16 and gradient checkpointing enabled. While SFT involves a simple forward pass, loss computation, and backward pass per step, both PPO and GRPO add extra steps such as LLM sampling, reward calculation, and advantage estimation. 

As detailed in Table \ref{tab:runtime}, both PPO and GRPO deliver notable performance improvements over SFT. Notably, PPO only incurs about a 4\% increase in per-step runtime compared to SFT. However, GRPO’s need to generate multiple responses per sample results in a higher runtime, despite its memory efficiency. Overall, our analysis highlights the trade-offs between these RL algorithms: PPO offers efficient runtime with the cost of additional overhead from the critic model, while GRPO reduces memory usage at the expense of increased sampling time.}

\subsection{Evaluating Zero-Shot Generalization of PPO Fine-Tuned Models and Comparison with GPT-4o}

\renewcommand\arraystretch{1.2}
\begin{table*}[!htb]
\small
\centering
\begin{tabular}{lccc}
  \toprule[1.2pt]
  \textbf{Tasks} & \textbf{LLAMA2-7B PPO-ST} & \textbf{LLAMA2-13B PPO-ST} & \textbf{GPT-4o} \\
  \midrule[1.2pt]
  \textbf{Sentiment Analysis} & & & \\
  \textit{Financial PhraseBank} & 97.2 & \textbf{97.7} & 97.5 \\
  \textit{Labelled Financial News} & 70.2 & \textbf{72.3} & 67.8 \\
  \textit{Mental Health} & \textbf{67.2} & 66.6 & 59.9 \\
  \textit{Emotion} & \textbf{78.0} & 76.4 & 77.6 \\
  \midrule[1.2pt]
  \textbf{Natural Language Inference} & & & \\
  \textit{Babi-nli} & 68.3 & \textbf{69.4} & 63.2 \\
  \textit{SIGA-nli} & 46.2 & \textbf{46.3} & 35.4 \\
  \hline 
  \textbf{Average} & {71.2} (4.3\textcolor{green}{$\uparrow$}) & \textbf{71.5} (4.6\textcolor{green}{$\uparrow$}) & {66.9} \\
  \bottomrule[1.2pt]
\end{tabular}

\caption{
Accuracy of different models across downstream tasks. For sentiment analysis tasks, models are fine-tuned on SST-2 and zero-shot evaluated on Financial PhraseBank~\citep{Malo2014GoodDO}, Labelled Financial News~\citep{arav_sood_2024}, Mental Health~\citep{huggingface-joangaes-depression}, and Emotion~\citep{saravia-etal-2018-carer}. Similarly, for natural language inference tasks, models are fine-tuned on MNLI and zero-shot evaluated on Babi-nli~\citep{weston2015towards} and SIGA-nli~\citep{nizamani-et-al-2024-siga}. PPO-ST represents fine-tuning using Proximal Policy Optimization. Gains over GPT-4o model in the average row is indicated with green arrows.
}
\label{tab:downstream}
\vspace{-0.3cm}
\end{table*}

We evaluate the zero-shot generalization capabilities of LLAMA2 7B and 13B models fine-tuned using PPO on a single dataset and subsequently tested across multiple other datasets (Table~\ref{tab:downstream}). For sentiment analysis tasks, the models were fine-tuned on SST-2 and evaluated on diverse datasets, including Financial PhraseBank~\citep{Malo2014GoodDO}, Labelled Financial News~\citep{arav_sood_2024}, Mental Health~\citep{huggingface-joangaes-depression}, and Emotion~\citep{saravia-etal-2018-carer}. Similarly, for natural language inference (NLI) tasks, the models were fine-tuned on MNLI and evaluated on Babi-nli~\citep{weston2015towards} and SIGA-nli~\citep{nizamani-et-al-2024-siga}.

Our results demonstrate that PPO fine-tuning improves the zero-shot performance of LLAMA2-chat-hf models compared to \textbf{GPT-4o}, a strong baseline. For sentiment analysis, \textcolor{black}{LLAMA2-13B-chat-hf} achieves 97.7\% accuracy on Financial PhraseBank, slightly outperforming GPT-4o (97.5\%). On Labelled Financial News, \textcolor{black}{LLAMA2-13B-chat-hf} records 72.3\%, exceeding GPT-4o by 4.5\%. Similarly, on the Mental Health dataset, \textcolor{black}{LLAMA2-7B-chat-hf} achieves 67.2\%, marking a notable gain of 7.3\% over GPT-4o. For the Emotion dataset, \textcolor{black}{LLAMA2-7B-chat-hf} achieves 78.0\%, with a smaller gain of 0.4\%. For NLI tasks, \textcolor{black}{LLAMA2-13B-chat-hf} achieves 69.4\% accuracy on Babi-nli, surpassing GPT-4o by 6.2\%. Additionally, \textcolor{black}{LLAMA2-13B-chat-hf} achieves 46.3\% accuracy on SIGA-nli, outperforming GPT-4o by more than 10\%. On average, both 7B and 13B versions of PPO fine-tuned LLAMA2-chat-hf models demonstrate a performance gain of over 4\% compared to GPT-4o, which is significantly larger in size and highly optimized.

\textcolor{black}{To ensure robust comparisons, we quantify uncertainty in our evaluations by generating 100 predictions for each example in the dataset. The evaluation metric is then computed over the entire dataset for each set, yielding a distribution of values. The 95\% confidence interval is defined by the 2.5th and 97.5th percentiles of this distribution. Results are presented in Table~\ref{tab:95confidence}.} 

These results demonstrate the effectiveness of simple PPO fine-tuning on a single task-specific dataset in significantly enhancing model performance on similar tasks. LLAMA2-chat-hf models fine-tuned with PPO consistently outperform GPT-4o across diverse downstream tasks, reinforcing PPO fine-tuning as a robust approach for improving the NLU capabilities of LLMs.

\textcolor{black}{We measured inference time on the Financial PhraseBank dataset with a batch size of 4. The BERT-base model, with 110M parameters, required 0.035s per step, while the LLAMA2-7B model, with 7B parameters and multi-token generation, took 0.997s per step. This difference is expected given the larger model size and the need for multiple forward passes in LLAMA2-7B. While LLM inference is slower, our focus is on improving natural language understanding with PPO, which achieves strong performance gains on both in-distribution and out-of-distribution NLU and NLI tasks.}

\subsection{\textcolor{black}{Evaluation of Instruction-Following in Out-of-Distribution Tasks}}

{\color{black}

To assess the instruction-following capabilities of LLMs in tasks differing from their fine-tuned format, we conduct evaluations using the LLAMA2-7B-chat-hf model fine-tuned on the SST-2 dataset. Specifically, we evaluate the performance of this model on the Amazon review task, which requires generating an integer rating between 1 and 5 based on the provided textual review. Although SST-2 and Amazon reviews both involve sentiment analysis, the two tasks differ distinctly in their input-output formatting, providing a clear measure of instruction-following adaptability.

We compare zero-shot prompting of three versions of the LLAMA2-7B-chat-hf model: the original non-fine-tuned model, a version fine-tuned using SFT, and another fine-tuned with PPO. The 95\% confidence intervals (CI) reported here are defined by the 2.5th and 97.5th percentiles of the bootstrap distribution. Using a consistent prompt template across models, we find that the PPO-fine-tuned model achieves an accuracy of 39.35\% (95\% CI: 38.39, 40.29), significantly outperforming the original model, which achieves 27\% accuracy (95\% CI: 19.00, 36.03). Conversely, the SFT-fine-tuned model demonstrates extremely poor performance, achieving less than 1\% accuracy.

\renewcommand\arraystretch{1.2}
\begin{table}[!htb]
\small
\centering
{\color{black}
\begin{tabular}{lcc}
  \toprule[1.2pt]
  \textbf{Method} & \textbf{Accuracy} & \textbf{95\% CI} \\
  \midrule[1.2pt]
  Non-fine-tuned & 27.00 & (19.00, 36.03) \\
  SFT      & 0.00961 & (0.00, 0.03) \\
  PPO      & \textbf{39.35} & (38.39, 40.29) \\
  \bottomrule[1.2pt]
\end{tabular}
}
\caption{\textcolor{black}{Performance of LLAMA2-7B-chat-hf on the Amazon Review dataset. Best results are highlighted in bold.}}
\label{amazon_review}
\end{table}

Qualitative analysis of sampled outputs reveals that the PPO-fine-tuned model reliably adheres to the instruction format and generates detailed reasoning to support its predictions. In contrast, the SFT-fine-tuned model often fails to adapt its responses to the required format, demonstrating limited generalization capabilities. PPO fine-tuning maintains proximity to the original model distribution through its clipping mechanism and the KL-divergence minimization term in the objective function (part of the TRL library), thereby preserving and enhancing the model’s intrinsic instruction-following capabilities. In contrast, SFT fine-tuning appears to narrow the model's learned distribution to task-specific training data, negatively impacting its original instruction-following proficiency.
}

\subsection{\textcolor{black}{Impact of Fine-Tuning on Language Modeling Ability}}
\label{impactonLM}
{\color{black}

We experiment with SFT and PPO to improve NLU capabilities of LLMs and observe improved performance using PPO. However, it is crucial to ensure that fine-tuning methods do not significantly degrade the models' general language generation abilities. To assess this, we directly evaluate the PPL~\citep{jelinek1977perplexity, Chelba2000StructuredLM} of LLAMA2-7B-chat-hf models fine-tuned on the SST-2 dataset using the WikiText-2 test set~\citep{merity2016pointer}, which follows a natural human-written text distribution. We compare these fine-tuned models against the original, non-fine-tuned baseline model, with the expectation that the PPL of the fine-tuned models should closely match the baseline. Our results reveal that the original LLAMA2-7B-chat-hf achieves a perplexity of 6.939. The PPO-fine-tuned model closely maintains this baseline performance with a perplexity of 6.966, indicating minimal impact on its general language modeling capabilities. 

In contrast, the SFT-fine-tuned model displays a higher perplexity of 7.384, suggesting a significant reduction in generation capabilities due to convergence toward task-specific training distributions. We conjecture that PPO's clipping mechanism and the KL-divergence minimization term in the objective function (part of the TRL library), effectively constrains policy updates, preventing large deviations from the reference model and thereby preserving the original language modeling capabilities of LLMs. These findings underscore PPO's effectiveness in maintaining the general language abilities of LLMs during fine-tuning.

\renewcommand\arraystretch{1.2}
\begin{table}[!htbp]
\small
\centering
{\color{black}
\begin{tabular}{lcc}
  \toprule[1.2pt]
  \textbf{Method} & \textbf{perplexity} \\
  \midrule[1.2pt]
  Non-fine-tuned & 6.939 \\
  SFT      & 7.384 \\
  PPO      & 6.966 \\
  \bottomrule[1.2pt]
\end{tabular}
}
\vspace{-0.1cm}
\caption{\textcolor{black}{Perplexity of LLAMA2-7B-chat-hf on the WikiText-2 test set. Lower perplexity indicates better language modeling ability.}}
\label{ppl}
\end{table}

}

\subsection{\textcolor{black}{Integrating a Reward Model}}
\label{rewardfunction}

\renewcommand\arraystretch{1.2}
\begin{table*}[!htbp]
\centering
\begin{subtable}[t]{0.48\textwidth}
\centering
\footnotesize
{\color{black}
\begin{tabular}{lc}
  \toprule[1.2pt]
  \textbf{Method} & \textbf{Accuracy (\%)} \\
  \midrule[1.2pt]
  PPO    & \textbf{96.4} \\
  PPO-RM & 89.7 \\
  \bottomrule[1.2pt]
\end{tabular}
}
\vspace{-0.1cm}
\caption{\textcolor{black}{SST-2 performance on GLUE.}}
\label{subtable:sst2p}
\end{subtable}
\hfill
\begin{subtable}[t]{0.48\textwidth}
\centering
\footnotesize
{\color{black}
\begin{tabular}{lc}
  \toprule[1.2pt]
  \textbf{Method} & \textbf{GPT Eval. Score} \\
  \midrule[1.2pt]
  PPO    & 3.479 \\
  PPO-RM & \textbf{4.104} \\
  \bottomrule[1.2pt]
\end{tabular}
}
\vspace{-0.1cm}
\caption{\textcolor{black}{Quality of generated analyses.}}
\label{subtable:gpt}
\end{subtable}
\vspace{-0.1cm}
\caption{\textcolor{black}{Comparison of reward function designs for LLAMA2-7B-chat-hf. The model trained with a rule based reward (PPO) achieves a high SST-2 classification accuracy of 96.4\%, while incorporating a sophisticated reward model (PPO-RM) significantly reduces accuracy (89.7\%) but yields substantially improved analysis quality, with a GPT evaluation score of 4.104 compared to 3.479 for the simple reward. Best results are highlighted in bold.}}
\label{combined_table-rm}
\end{table*}

{\color{black}
While our primary reward function is based on matching generated outputs to true labels, we recognize that more sophisticated reward designs may be necessary for complex NLU tasks. To address this, we investigate the effect of integrating a reward model into our PPO training, with the aim of enhancing not only classification performance on SST-2 but also the quality of generated analyses.

\textbf{Reward Modeling Setup.} For the first 5,000 training samples of the SST-2 dataset, LLAMA2-7B-chat-hf generates four responses per data point. Each response includes a sentiment judgment (Positive/Negative) and a supporting analysis. To robustly rank these responses, we use GPT-4o as an evaluator. GPT-4o ranks the responses based on: (i) the correctness of the sentiment judgment (i.e., matching the ground truth), (ii) the consistency between the judgment and its accompanying analysis, and (iii) the overall factual correctness and helpfulness of the analysis. To ensure clear differentiation, we include two reference responses—one with only the correct answer and one with only the incorrect answer—and define the ranking order as: correct answer with analysis > only correct answer > incorrect answer with analysis > only incorrect answer.

\textbf{Training the Reward Model.} A reward model is then trained on this ranked dataset using a BERT-based architecture (bert-base-cased). For each input \(x\), we consider pairs of responses \((y_w, y_l)\), where \(y_w\) denotes a response ranked higher by our evaluator (GPT-4o) due to its correct sentiment and coherent analysis, and \(y_l\) denotes a lower-ranked response. The model learns to assign higher scores to better responses via a pairwise ranking loss:
\begin{equation}
\begin{split}
L(\theta) 
  &= -\mathbb{E}_{(x,y_w,y_l)\sim D} \\
  &\quad\Bigl[\log \sigma\bigl(r_\theta(x,y_w) - r_\theta(x,y_l)\bigr)\Bigr],
\end{split}
\end{equation}
where \(r_\theta(x,y)\) is the score assigned to response \(y\) given \(x\), and \(\sigma\) is the sigmoid function converting the score difference into a probability. This loss encourages the reward model to output higher scores for responses with superior judgments and analyses.

\textbf{Incorporating the Reward Model into PPO Training.} During PPO training on SST-2, LLM is tasked with generating both a sentiment judgment and an analysis. The trained reward model provides the reward signal by scoring these outputs. As shown in Table~\ref{subtable:sst2p}, while the PPO model trained with reward signals from the reward model (PPO-RM) produces analyses of higher quality, it suffers from a significant reduction in classification performance, dropping from 96.4\% to 89.7\%. We believe this discrepancy might be due to the limited sample size used for reward model training and potential reward hacking~\cite{amodei2016concrete} during optimization. However, we will explore this further in our future works.

\textbf{Evaluation of Generated Analyses.} To further assess the impact of our reward design, we evaluated the quality of generated analyses. We sampled 100 data points from three models: the original LLAMA2-7B-chat-hf, the PPO model trained using only correct-answer rewards (PPO), and the PPO model trained with the reward model (PPO-RM). GPT-4o then scored each analysis on a scale from 1 to 5 based on answer correctness and logical coherence. As indicated in Table~\ref{subtable:gpt}, the PPO model using reward model signals achieved the highest average score, suggesting that a more complex reward function can enhance the quality of generated outputs.

In summary, while the integration of a reward model in PPO training significantly reduces classification performance compared to using only correct-answer rewards, it considerably improves the GPT evaluation scores of the analyses produced by the LLM. 
}

\section{Conclusion}
Prompting-based approaches, including zero-shot and few-shot prompting, are commonly used to adapt LLMs to downstream tasks. However, our experiments show that when applied to \textcolor{black}{LLAMA2-7B-chat-hf}, these methods underperform on NLU benchmarks such as GLUE and SuperGLUE (table~\ref{SuperGLUE_result}), often trailing smaller encoder-only models like BERT-base. To address this, we investigate two fine-tuning strategies that update only LoRA layers for computational efficiency: SFT and PPO. While SFT yields modest improvements, PPO provides substantial gains by framing NLU tasks as reinforcement learning problems. PPO-tuned models not only outperform strong baselines like BERT-large but also generalize well across model families and tasks. Notably, PPO-trained \textcolor{black}{LLAMA2-7B-chat-hf} outperforms GPT-4o by 10.8\% on SIGA-nli and 7.3\% on the Mental Health dataset, demonstrating strong zero-shot generalization from single-task fine-tuning. More broadly, we highlight a promising direction: adapting LLMs to new tasks without labeled data by using reward-driven learning. With a well-defined reward function, PPO can steer models toward high-reward behaviors—offering a scalable, label-efficient alternative to SFT.

\section*{Limitations}

This work takes an initial step toward framing NLU as a reinforcement learning problem for LLMs under 14B parameters. While our long-term goal is to reduce reliance on curated datasets by leveraging richer, task-specific reward models, we currently adopt a simple binary reward signal based on exact label matching. This design enables a controlled evaluation of PPO as an effective adaptation strategy, showing consistent gains over prompting and supervised fine-tuning. Although we present a preliminary exploration of model-driven reward functions in Section~\ref{rewardfunction}, further research is needed to develop robust and generalizable reward signals that can support more complex or weakly supervised tasks without requiring extensive manual annotation. Overall, our findings suggest that casting nuanced tasks as reinforcement learning problems, through the design of appropriate environments and reward functions, offers a scalable and flexible alternative to standard fine-tuning, particularly when the model is already well-initialized through pretraining.

\bibliography{custom}

\appendix

\newpage
\appendix

\section{Related Works}
\label{sec: related-works}

Policy-based reinforcement learning (RL) directly optimizes an agent’s policy by learning its parameters to maximize long-term rewards. Unlike value-based methods like Q-learning~\citep{watkins1992q} and DQN~\citep{hester2018deep}, which indirectly derive policies through value functions, policy-based methods represent the policy as a parameterized function. This function, $p_{\theta}(a|s)$, defines the probability of taking action $a$ in state $s$, where $\theta$ represents the policy parameters. The goal is to learn optimal parameters $\theta^*$ that maximize the expected cumulative reward, typically through policy gradient methods~\citep{sutton1999policy}. These methods excel in high-dimensional or continuous action spaces, where value-based methods can struggle~\citep{deisenroth2013survey}.

Policy-based methods in reinforcement learning (RL) have evolved significantly over time, starting with REINFORCE~\citep{williams1992simple}, which optimizes policies using the policy gradient theorem but suffers from high variance due to its reliance on Monte Carlo estimates of the reward. Monte Carlo estimates refer to calculating the total reward based on full episodes of interaction, meaning updates are made only after an entire sequence of actions and rewards is observed, which can lead to noisy and slow learning. To address this, actor-critic methods like A2C and A3C~\citep{mnih2016asynchronous} introduced a critic that estimates the value of the current state, allowing for smoother updates by reducing the variability in policy updates and speeding up convergence. However, these methods still faced instability when large updates caused the new policy to diverge too far from the previous one. Trust Region Policy Optimization (TRPO)~\citep{schulman2015trust} tackled this by limiting the size of policy updates using a KL divergence constraint, but its implementation was complex and computationally expensive. Proximal policy optimization (PPO)~\citep{schulman2017proximal} simplified this process by introducing a clipped objective function that keeps policy updates within a stable range while being easier to implement. PPO's balance between simplicity and stability has made it a widely adopted method in modern RL research.

In NLP, PPO has been effectively used in reinforcement learning from human feedback (RLHF) to align LLM outputs with human preferences, as seen in works like InstructGPT~\citep{ouyang2022training} and Constitutional AI~\citep{bai2022constitutional}. These approaches treat the LLM as a policy, where model responses are actions, and human feedback serves as rewards. PPO updates the policy based on the reward model trained on human preferences. Additionally, policy-based RL methods have been applied to enhance LLM reasoning capabilities~\citep{ziegler2019fine, havrilla2024teaching, hu2023language}. In this work, we apply PPO to fine-tune LLMs on NLU tasks.

\section{Preliminaries on Application of PPO for Fine-tuning LLMs}
\label{sec:preliminaries}

Proximal policy optimization (PPO)~\citep{schulman2017proximalpolicyoptimizationalgorithms} is an online reinforcement learning algorithm. In this section, we describe the process to fine-tune an LLM using PPO. During training, at each timestep \(t\), the LLM (policy) generates a token prediction \(a_t\) (action) based on the state \(s_t\), which consists of the sequence of generated tokens up to timestep \(t-1\). The final generated output is evaluated in the context of the downstream task, where the environment provides feedback in the form of rewards. The model updates its parameters based on these rewards to improve its ability to generate accurate predictions over time.

\paragraph{Actor Model}

PPO uses gradient ascent to optimize the following objective, aiming to maximize cumulative rewards:
\begin{equation}
\begin{split}
J(\theta) 
&= \mathbb{E}_{(s_t,a_t)\sim\pi_{\theta'}}\Biggl[ 
    \min\Bigl(\,
      \frac{p_{\theta}(a_t\mid s_t)}{p_{\theta'}(a_t\mid s_t)}\,\hat A_t,\\
&\qquad\quad
      \mathrm{clip}\!\bigl(
        \tfrac{p_{\theta}(a_t\mid s_t)}{p_{\theta'}(a_t\mid s_t)},\,1-\epsilon,\,1+\epsilon
      \bigr)\hat A_t
    \Bigr)
  \Biggr]
\end{split}
\label{eq:ppo_objective}
\end{equation}
Here, \( p_{\theta}(a_t | s_t) \) is the probability of taking action \( a_t \) in state \( s_t \) under the current policy, while \( p_{\theta'}(a_t | s_t) \) represents this probability under the old policy. In PPO, the training data—specifically, the state-action pairs \((s_t, a_t)\)—are sampled using the old policy \(\pi_{\theta'}\) (the LLM before it is updated), rather than the new policy currently being optimized. Thus, the ratio \( \frac{p_{\theta}(a_t | s_t)}{p_{\theta'}(a_t | s_t)} \) accounts for how much the new policy has changed relative to the old policy and adjusts the likelihood of an action accordingly. This ratio is multiplied by \( \hat{A}_t \), the Generalized Advantage Estimation (GAE)~\citep{schulman2018highdimensionalcontinuouscontrolusing}, which measures how much \textit{better or worse} an action \( a_t \) is compared to the expected outcome under the current policy.
\[
    \hat{A}_t = R_t + \gamma V_{t+1} - V_t + \gamma \lambda \hat{A}_{t+1},
\]
Here, \( R_t + \gamma V_{t+1} - V_t \) represents the temporal difference (TD) error~\citep{sutton1988learning}. In this expression, \( R_t \) is the immediate reward received after taking action \( a_t \), \( V_t \) is the expected reward before the action, and \( \gamma V_{t+1} \) is the discounted estimate of the future reward after the action. This term reflects how the action $a_t$ performed when compared to the expected return at state $s_t$. The second term, \( \gamma \lambda \hat{A}_{t+1} \), is the smoothing factor in GAE, where \( \lambda \) is the trade-off parameter. This recursive estimate allows the model to incorporate future information, making the advantage estimate more stable. Smaller values of \( \lambda \) emphasize on immediate rewards, while larger values capture longer-term dependencies. The discount factor \( \gamma \) controls how much emphasis is placed on future rewards compared to immediate ones, with higher values of \( \gamma \) giving more weight to future rewards. \( V_t \), which represents the expected future reward from state \( s_t \), is estimated by a \textit{critic model}. 

The clipping function \( \text{clip}(\text{ratio}, 1-\epsilon, 1+\epsilon) \) limits the change between the current and old policy, ensuring stable updates by preventing large deviations. This helps avoid too-large policy changes that could destabilize training. In summary, PPO optimizes the policy using gradient ascent to maximize cumulative rewards while ensuring stable updates through clipping, with the GAE providing a more stable and accurate advantage estimate by incorporating future information recursively.

\paragraph{Critic Model}
The critic model consists of a value head, which is a multi-layer perceptron attached to the final layer of the LLM. It takes the LLMs representation of the generated token sequence up to timestep \(t\) (i.e., the state \(s_t\)) and predicts a scalar value representing the value function \(V_t\) for that state. The critic model is updated using the square of TD error, which is computed as:
\begin{equation}
    \delta_t = (R_t + \gamma V_{t+1} - V_t)^2,
\end{equation}
where \( \delta_t \) represents the L-2 loss between the actual reward \( R_t \), combined with the discounted estimate of future rewards \( \gamma V_{t+1} \), and the current predicted value \( V_t \) for state \( s_t \). By minimizing this TD error via gradient descent, the critic model updates its value function predictions, improving alignment with the actual rewards and future outcomes. In summary, LLM uses the PPO objective to update its policy based on feedback from the critic model, while the critic model is updated to better predict the value function for future states.

We implement PPO using TRL library~\cite{huggingface_trl_docs} that includes a KL divergence term in the PPO objective~\ref{eq:ppo_objective}.

\section{Hyperparameter Settings}
\label{sec:hyperparameters}

For PPO-based fine-tuning, grid search is performed to select the batch size in 4, 8, 12, and 16 for each task. A batch size of 24 was used across all tasks during supervised fine-tuning (SFT). The PPO epoch is set to 4, meaning that each sampled batch is used for updating the model four times. The initial learning rate for all tasks was set to \(9 \times 10^{-6}\). We utilized the Adafactor optimizer for PPO training and AdamW for SFT. A cosine annealing learning rate scheduler with a warmup phase was employed, where the learning rate was gradually increased during the first 10\% of training steps and then reduced to one-tenth of the initial value by the end of training. We use a rank $r=16$ for the LoRA layers. We trained both PPO and SFT models until convergence on the validation set. The best hyperparameters were selected based on performance on the validation set. The final reported results for the GLUE and SuperGLUE are from their corresponding evaluation server. For evaluation, multinomial sampling with a temperature of 1 was used to generate a single response per data sample. The model generated responses with lengths between 12 and 32 tokens, with the generation process concluding using a special identifier \texttt{``</Judgement>''}. 

For few-shot evaluations, we conducted a sweep over 1–5 shots on the validation set and reported results using the best-performing configuration for each experiment. We ensured that the total input length, including few-shot demonstrations, remained within the model’s context window (e.g., in MultiRC, where input lengths can be long). For our experiments, we utilized the TRL library ~\citep{vonwerra2022trl} on a single Nvidia A100 GPU.

\renewcommand\arraystretch{1.2}
\begin{table*}[!htb]
\centering
\small
\begin{tabular}{lccc}
  \toprule[1.2pt]
  \textbf{Tasks} & \textbf{LLAMA2-7B PPO-ST} & \textbf{LLAMA2-13B PPO-ST} & \textbf{GPT-4o} \\
  \midrule[1.2pt]
  \textbf{Sentiment Analysis} & & & \\
  \textit{Financial PhraseBank} & \textcolor{black}{(96.2, 98.1)} & {\textcolor{black}{(96.9, 98.5)}} & \textcolor{black}{(96.6, 98.4)} \\
  \textit{Labelled Financial News} & \textcolor{black}{(66.1, 74.6)} & {\textcolor{black}{(69.0, 76.6)}} & \textcolor{black}{(63.2, 72.2)} \\
  \textit{Mental Health} & {\textcolor{black}{(66.6, 67.7)}} & \textcolor{black}{(66.0, 67.1)} & \textcolor{black}{(59.3, 60.5)} \\
  \textit{Emotion} & {\textcolor{black}{(77.4, 78.6)}} & \textcolor{black}{(75.8, 77.0)} & \textcolor{black}{(77.0, 78.2)} \\
  \midrule[1.2pt]
  \textbf{Natural Language Inference} & & & \\
  \textit{Babi-nli} & \textcolor{black}{(64.3, 71.5)} & {\textcolor{black}{(65.1, 73.0)}} & \textcolor{black}{(58.8, 67.6)} \\
  \textit{SIGA-nli} & {\textcolor{black}{(39.0, 53.9)}} & \textcolor{black}{(40.6, 53.7)} & \textcolor{black}{(28.5, 42.2)} \\
  \bottomrule[1.2pt]
\end{tabular}
\caption{\textcolor{black}{
To quantify uncertainty in our evaluations, we generate 100 predictions for each example in the dataset. The evaluation metric is then computed for each set over the entire dataset, forming a distribution of values. The 95\% confidence interval is defined by the 2.5th and 97.5th percentiles of this distribution. For sentiment analysis, models fine-tuned on SST-2 are evaluated in a zero-shot setting on Financial PhraseBank, Labelled Financial News, Mental Health, and Emotion datasets. For natural language inference, models fine-tuned on MNLI are zero-shot evaluated on Babi-NLI and SIGA-NLI.
}}
\label{tab:95confidence}
\end{table*}

\section{Reward Curve for PPO Fine-Tuning}
\label{sec:curve}

\begin{figure*}[ht]
    \centering
    \includegraphics[width=0.8\textwidth]{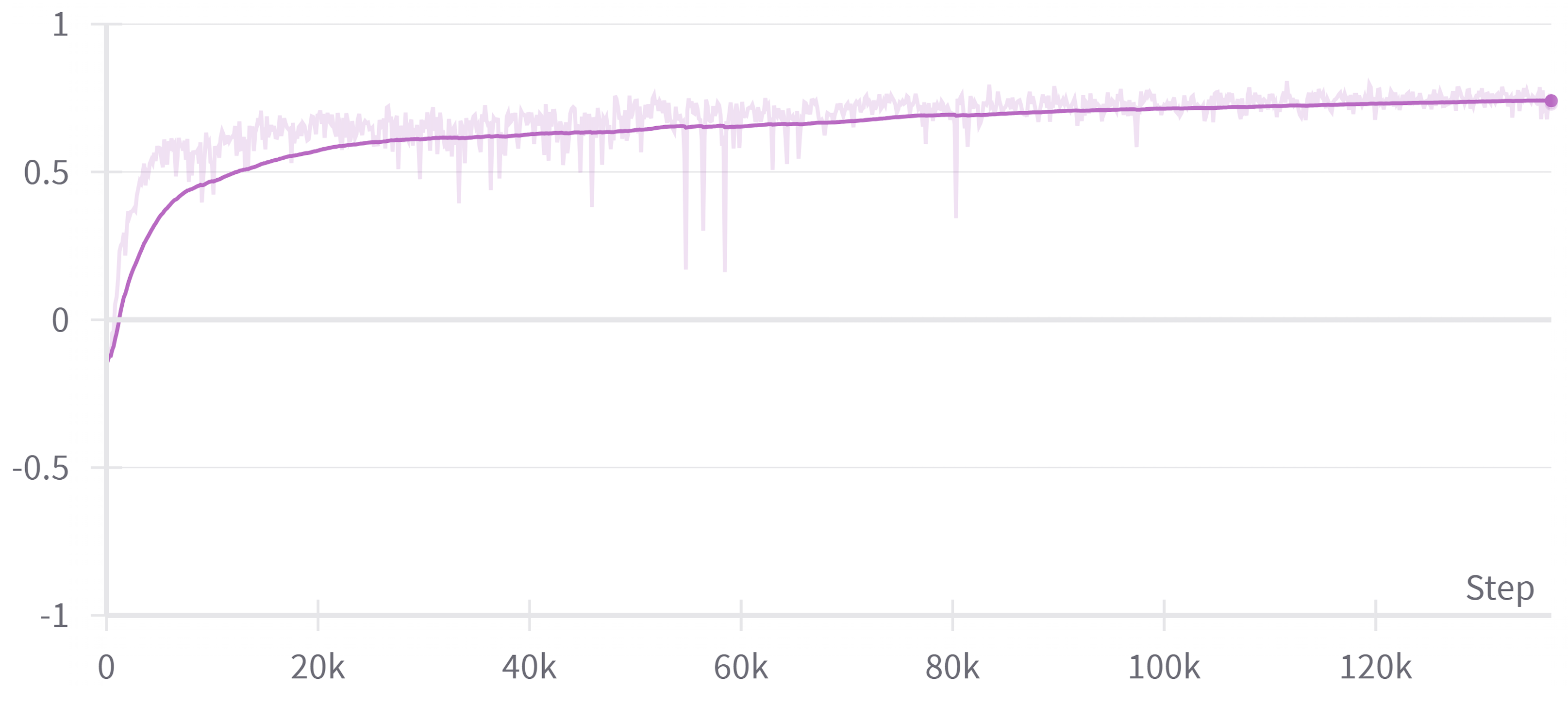}
    \caption{Reward curve for multitask PPO fine-tuning of \textcolor{black}{LLAMA2-7B-chat-hf} on the GLUE dataset. The plot illustrates the relationship between training iterations (x-axis) and reward values (y-axis), demonstrating the effectiveness of the PPO optimization approach in improving model performance over time.}
    \label{fig:reward_curve}
\end{figure*}

We present the reward curve from fine-tuning \textcolor{black}{LLAMA2-7B-chat-hf} using PPO in a multitask setting on the GLUE dataset. Figure \ref{fig:reward_curve} illustrates the reward values over training iterations, offering insights into the training dynamics of the model. The curve serves as a key performance metric, tracking the model's learning progress across multiple tasks. The consistent upward trend demonstrates that PPO fine-tuning effectively improves \textcolor{black}{LLAMA2-7B-chat-hf}’s ability to generate task-relevant outputs.

\section{Confidence Intervals on Validation Set of GLUE Benchmark}
\label{glue_confidence}

Our GLUE results are obtained from the official online evaluation servers, which impose a submission limit and keep test-set labels hidden, so applying a bootstrapping procedure directly is not feasible. We compute 95\% confidence intervals on the validation sets of the GLUE benchmark (available offline) using the following procedure (used in the paper). For each example in a given dataset, we generate 100 independent predictions. We then compute the evaluation metric over the entire dataset for each set of predictions, yielding a distribution of metric values. The 2.5th and 97.5th percentiles of this distribution define the bounds of our 95\% confidence interval. The resulting intervals are reported in the table~\ref{tab:glue_confidence}.

\renewcommand\arraystretch{1.2}
\begin{table*}[!htb]
\centering
\small
\begin{tabular}{lcc|cc}
  \toprule[1.2pt]
  \textbf{Dataset} & \textbf{PPO (Eval Metric)} & \textbf{PPO (95\% CI)} & \textbf{SFT (Eval Metric)} & \textbf{SFT (95\% CI)} \\
  \midrule[1.2pt]
  {CoLA} & 64.54 & (59.24, 70.08) & 60.10 & (54.79, 65.13) \\
  {MRPC} & 88.41 & (86.92, 91.00) & 85.28 & (84.24, 87.19) \\
  {STS-B} & 88.57 & (87.07, 89.98) & 85.68 & (83.85, 86.72) \\
  {QQP} & 83.34 & (82.83, 83.73) & 79.76 & (79.22, 80.17) \\
  {QNLI} & 93.17 & (92.50, 93.88) & 93.67 & (92.90, 94.35) \\
  {MNLI-matched} & 88.63 & (88.05, 89.30) & 87.02 & (86.32, 87.64) \\
  {MNLI-unmatched} & 88.95 & (88.28, 89.53) & 87.27 & (86.54, 87.89) \\
  {RTE} & 84.90 & (80.26, 89.64) & 81.95 & (77.56, 86.95) \\
  {SST-2} & 96.10 & (94.56, 97.54) & 96.44 & (94.95, 97.37) \\
  {WNLI} & 74.65 & (63.30, 85.87) & 66.20 & (49.30, 71.83) \\
  \bottomrule[1.2pt]
\end{tabular}
\caption{\textcolor{black}{
Validation set results on GLUE benchmark. Metrics: CoLA (Matthews corr.), MRPC \& QQP (F1), STS-B (Spearman corr.), all others (accuracy). 95\% confidence intervals are shown in parentheses.
}}
\label{tab:glue_confidence}
\end{table*}

\section{Results on SuperGLUE Benchmark}
\label{superglue_benchmark}

\renewcommand\arraystretch{1.2}
\begin{table*}[!htbp]
\centering
\small
\begin{tabular}{lcccccc}
  \toprule[1.2pt]
  \textbf{Models} & \textbf{BoolQ} & \textbf{CB} & \textbf{COPA} & \textbf{MultiRC} & \textbf{ReCoRD} & \textbf{RTE} \\
  \midrule[1.2pt]
  \textbf{BERT-large} & 77.4 & \underline{75.7}/83.6 & 70.6 & 70.0/24.0 & \textbf{72.0/71.3} & 71.6 \\
  \textbf{BERT-large++} & \underline{79.0} & \textbf{84.7}/90.4 & \underline{73.8} & \underline{70.0}/24.1 & \textbf{72.0/71.3} & \underline{79.0} \\
  \multicolumn{1}{l}{\textbf{\textcolor{black}{LLAMA2-7B-chat-hf}}} & & & & & & \\
  \textit{Zero-shot prompting} & 75.8 & 26.4/43.6 & 57.0 & 51.9/20.3 & 27.0/26.2 & 59.2 \\
  \textit{Few-shot prompting} & 80.2 & \textcolor{black}{49.8/66.0} & \textcolor{black}{73.4} & \textcolor{black}{46.6/15.4} & \textcolor{black}{36.3/35.3} & 72.9 \\
  \textit{PPO-ST} & \textbf{85.9} & 74.7/\underline{88.0} & \textbf{88.6} & \textbf{82.5}/50.0 & 70.6/69.9 & \textbf{84.3} \\
  \bottomrule[1.2pt]
\end{tabular}

\vspace{0.15cm} 

\small
\begin{tabular}{lcccccc}
  \toprule[1.2pt]
  \textbf{Models} & \textbf{WiC} & \textbf{WSC} & \textbf{AXb} & \textbf{AXg} & \textbf{Average} \\
  \midrule[1.2pt]
  \textbf{BERT-large} & \underline{69.5} & \underline{64.3} & 23.0 & \underline{97.8}/51.7 & 69.0 \\
  \textbf{BERT-large++} & \underline{69.5} & \underline{64.3} & \underline{38.0} & \textbf{99.4}/51.4 & \underline{71.5} \\
  \multicolumn{1}{l}{\textbf{\textcolor{black}{LLAMA2-7B-chat-hf}}} & & & & & \\
  \textit{Zero-shot prompting} & 54.4 & 52.1 & 9.1 & 64.0/55.1 & 49.5 \\
  \textit{Few-shot prompting} & \textcolor{black}{54.4} & \textcolor{black}{62.3} & 9.1 & 64.0/55.1 & \textcolor{black}{54.9} \\
  \textit{PPO-ST} & \textbf{72.1} & \textbf{78.1} & \textbf{52.7} & 91.0/\textbf{79.8} & \textbf{78.3} \\
  \bottomrule[1.2pt]
\end{tabular}
\vspace{-0.1cm}
\caption{SuperGLUE test results are scored by the evaluation server (\href{https://super.gluebenchmark.com/}{SuperGLUE benchmark}). The experimental data for BERT-large and BERT-large++ are taken from the original SuperGLUE paper~\citep{wang2020supergluestickierbenchmarkgeneralpurpose}.
The metrics used in the experiments are as follows: CB: F1 / Acc; MultiRC: F1 / Exact Match; ReCoRD: F1 / Exact Match; AXb: MCC; AXg: Gender parity score / Acc. For the remaining tasks not mentioned, accuracy (Acc) is reported. Average column corresponds to the averaged performance across all the datasets. For tasks with multiple evaluation metrics, we first compute the average of those metrics to obtain a single task score, which is then used in the overall average calculation. The \textbf{bolded} results indicate the best results, and the \underline{underlined} results indicate the second-best results.}
\label{SuperGLUE_result}
\end{table*}

We fine-tuned the \textcolor{black}{LLAMA2-7B-chat-hf} model using PPO on the SuperGLUE dataset and compared its performance against several baselines, including BERT-large, BERT-large++, and zero-shot and few-shot prompting of \textcolor{black}{LLAMA2-7B-chat-hf}. The term ``BERT++'' refers to a BERT model fine-tuned using the supplementary training on intermediate labeled-data tasks (STILTs) approach~\citep{phang2018sentence}, where the model is first fine-tuned on related transfer tasks before being fine-tuned on SuperGLUE tasks. For example, MNLI from the GLUE benchmark~\citep{wang2018glue} is used as an intermediate task for CB, RTE, and BoolQ~\citep{wang2020supergluestickierbenchmarkgeneralpurpose}. In contrast, our experiments with LLM did not use this method. Our models were only fine-tuned on the datasets in the SuperGLUE benchmark.

As shown in Table \ref{SuperGLUE_result}, the PPO-tuned \textcolor{black}{LLAMA2-7B-chat-hf} achieved the highest average performance, surpassing all baselines. PPO demonstrated particularly strong improvements on reasoning-intensive tasks like COPA and MultiRC, where it significantly outperformed both prompting methods and encoder-only models. These results highlight the effectiveness of PPO in improving the model’s capabilities, particularly for tasks requiring reasoning and contextual understanding.

\textcolor{black}{It is worth noting that on MultiRC, few-shot prompting performs slightly worse than zero-shot prompting. This may be because MultiRC involves long input contexts, and incorporating examples in a few-shot prompt can cause the total input length to approach or exceed the LLMs maximum context window. 
}

\section{\textcolor{black}{Evaluation on Reading Comprehension Tasks}}

{\color{black}
We evaluate LLAMA2-7B-chat-hf on the SQuAD reading comprehension task, where the objective is to select a passage from a given context that best answers a question. Two fine-tuning strategies are compared: SFT, which directly uses the ground-truth answer as the training label, and PPO, which leverages reward functions based on Exact Match (EM) and F1 Score. EM metric is computed by comparing a normalized prediction against the normalized ground truth (with normalization involving lowercasing and punctuation removal); a perfect match yields an EM score of 1, otherwise 0. F1 score measures word-level overlap, balancing how many predicted words are correct (precision) and how many ground-truth words are included (recall).

Models were fine-tuned for one epoch on the SQuAD training set and evaluated on the development set. In our evaluation, zero-shot prompting yields an EM of 7.66 and an F1 score of 32.27. SFT significantly improves these metrics (EM: 59.17, F1: 76.48), while PPO further enhances performance, achieving an EM of 65.74 and an F1 score of 81.82—corresponding to improvements of 6.57 and 5.34 points over SFT, respectively.

These results indicate that optimizing with reward functions based on EM and F1 via PPO leads to further improvements in reading comprehension performance, thereby validating our approach relative to both zero-shot prompting and standard SFT.
}

\renewcommand\arraystretch{1.2}
\begin{table}[!htbp]
\centering
{\color{black}
\begin{tabular}{lcc}
  \toprule[1.2pt]
  \textbf{Method} & \textbf{EM} & \textbf{F1} \\
  \midrule[1.2pt]
  Non-fine-tuned & 7.66 & 32.27 \\
  SFT      & 59.17 & 76.48 \\
  PPO      & \textbf{65.74} & \textbf{81.82} \\
  \bottomrule[1.2pt]
\end{tabular}
}
\vspace{-0.1cm}
\caption{\textcolor{black}{Performance of LLAMA2-7B-chat-hf on the SQuAD dataset. PPO uses Exact Match and F1 as reward signals. Best results are highlighted in bold.}}
\label{squad_results}
\end{table}

\begin{table}[!htbp]
\small
\centering
\begin{tabularx}{\linewidth}{lXX}
  \toprule[1.2pt]
  \textbf{Dataset} & \textbf{LLaMA2-7B-chat-hf} & \textbf{LLaMA2-7B-base} \\
  \midrule[1.2pt]
  CoLA           & \textbf{64.54} & 57.29 \\
  MRPC           & \textbf{88.41} & 80.35 \\
  STS-B          & \textbf{88.57} & 86.81 \\
  QQP            & 83.34 & \textbf{84.39} \\
  QNLI           & \textbf{93.17} & 92.95 \\
  MNLI-matched   & \textbf{88.63} & 87.15 \\
  MNLI-unmatched & \textbf{88.95} & 88.22 \\
  RTE            & \textbf{84.90} & 83.75 \\
  SST-2          & 96.10 & \textbf{96.56} \\
  WNLI           & \textbf{74.65} & 67.61 \\
  \midrule
  \textbf{Average} & \textbf{85.13} & 82.51 \\
  \bottomrule[1.2pt]
\end{tabularx}
\caption{Performance comparison of PPO fine-tuning on LLaMA2-7B-chat-hf and LLaMA2-7B-base across GLUE tasks. Best results for each dataset are highlighted in bold.}
\label{glue_results}
\end{table}

\section{Comparison of LLaMA2-7B-base and LLaMA2-7B-chat-hf on GLUE Tasks}

We fine-tuned LLaMA2-7B-base with PPO and compared it to PPO-tuned LLaMA2-7B-chat-hf on GLUE tasks. The instruction-tuned chat model, with enhanced prompt-following capabilities, outperforms the base model on most tasks, averaging 85.13 vs. 82.51. This highlights the benefit of starting from an instruction-tuned checkpoint for task-specific RL.

\end{document}